\documentclass[acmsmall]{acmart}

\usepackage{subcaption}
\AtBeginDocument{%
  \providecommand\BibTeX{{%
    \normalfont B\kern-0.5em{\scshape i\kern-0.25em b}\kern-0.8em\TeX}}}

\setcopyright{none}

\begin{document}


\title{WordBias: An Interactive Visual Tool for Discovering Intersectional Biases Encoded in Word Embeddings}


\author{Bhavya Ghai}
\affiliation{\institution{Stony Brook University}}
\email{bghai@cs.stonybrook.edu}

\author{Md Naimul Hoque}
\affiliation{\institution{Stony Brook University}}
\email{mdhoque@cs.stonybrook.edu}

\author{Klaus Mueller}
\affiliation{\institution{Stony Brook University}}
\email{mueller@cs.stonybrook.edu}

\renewcommand{\shortauthors}{Ghai, et al.}

\begin{abstract}
\textit{Intersectional bias} 
is a bias caused by an overlap of multiple social factors like gender, sexuality, race, disability, religion, etc.
A recent study has shown that word embedding models can be laden with biases against intersectional groups like African American females, etc. The first step towards tackling such intersectional biases is to identify them. 
However, discovering biases against different intersectional groups remains a challenging task.   
In this work, we present \textit{WordBias}, an interactive visual tool designed to explore biases against intersectional groups encoded in static word embeddings. Given a pretrained static word embedding, WordBias computes the association of each word along different groups based on race, age, etc. and then visualizes them using a novel interactive interface.
Using a case study, we demonstrate how WordBias can help uncover biases against intersectional groups like Black Muslim Males, Poor Females, etc. encoded in word embedding. 
In addition, we also evaluate our tool using qualitative feedback from expert interviews. 
\end{abstract}



\maketitle

\section{Introduction}
Word embedding models such as Glove \cite{glove} and Word2vec \cite{mikolov2013distributed} can be understood as a mapping between a word and its corresponding vector representation. They serve as the foundational unit for many NLP applications such as sentiment analysis, machine translation, etc. and could possibly be used to bootstrap any NLP task~\cite{surveyEval}. It has been shown that word embedding can learn and exhibit social biases based on race, gender, ethnicity, etc. that are encoded in the training dataset \cite{bolukbasi2016man, 100years, narayan2017semantics}. Social biases in word embeddings are manifested as stereotypes or undesirable associations between words \cite{100years}. For example, word embedding models 
might disproportionately associate Male names with career and math, while Female names might be associated with family and arts \cite{100years}. Existing literature has mostly focused on measuring and mitigating the \textit{individual} social biases based on race, gender, etc. encoded in word embeddings \cite{bolukbasi2016man, 100years, narayan2017semantics, vargas2020exploring, kumar2020nurse, guo2020detecting, dev2019attenuating}.

Recent studies have shown the presence of \textit{Intersectional Bias} 
in AI systems \cite{kim2020intersectional, genderShades, guo2020detecting} i.e. a bias towards a population defined by multiple sensitive attributes like `black muslim females' \cite{crenshaw2017intersectionality, intersectionality_book}. For example, facial recognition software applications have been shown to perform worse for the intersectional group `darker females' than for either darker individuals or females \cite{genderShades}. Similarly, word embedding models have also been shown to contain biases against intersectional groups like Mexican American females \cite{guo2020detecting}. When such biased word embeddings are used for any downstream application, their inherent social biases are propagated further, which can cause discrimination \cite{german,zhao2018gender}. Hence, it becomes critical to investigate the presence of different intersectional biases before using it for some application.

Stereotypes associated to an intersectional group say `Black Males' are composed of stereotypes pertaining to constituting subgroups (Blacks and Males) along with some unique elements \cite{ghavami2013intersectional}. The proportion of stereotypes which overlap with either of the constituting subgroups can vary based on the intersectional group. For example, a study on 627 undergraduate students found that the percentage of overlap for intersectional groups like White men is 81\%, White women is 88\%, Black women is 44\%, Black men is 73\%, Middle Eastern American men is 91\%, etc. \cite{ghavami2013intersectional}. This work focuses on this overlapping aspect of intersectionality. Given that word embedding models can consist of thousands of unique words and the number of intersectional groups can increase drastically with the number of sensitive attributes considered, it becomes challenging to explore the massive space of possible associations. Writing custom code to test the different associations can be tedious and ineffective. 
In this work, we present the first interactive visual tool, \textit{WordBias}, for exploring biases against different intersectional groups encoded in word embeddings. Given a pretrained word embedding, our tool computes the association (bias score) of each word along different social categorizations (bias types) like gender, religion, etc. and then visualizes them using a novel interactive interface (see \autoref{fig:teaser}). Here, each categorization (bias type) e.g. race consists of two subgroups, say Blacks and Whites. Using bias metrics, WordBias computes the degree to which a word aligns with one subgroup over the other. 
The visual interface then allows the user to \textit{investigate} how a specific word associates with different individual subgroups and also \textit{discover} words that are associated with an intersectional group. Considering the overlapping aspect of intersectionality, WordBias considers a word to be associated with an intersectional group say `Christian Males' if it associates strongly with each of its constituting subgroups (Christians and Males).


Users can interact with our tool to explore the space of word associations and then use their real world knowledge to determine if a given association is socially desirable. For example, the association between the word `queen' and female is desirable whereas the association between `teacher' and female is not. 
Using a case study, we demonstrate how WordBias can help discover biases against different intersectional groups like `Young Poor Blacks', `Black Muslim Males', etc. in Word2Vec embedding. Identifying such biases can serve as the first step toward deterring its spread and help develop counter-strategies. 
Lastly, we evaluate the usability and utility of our tool using qualitative feedback from domain experts. 
We have made the source code for our tool along with a live demo publicly available for easy reproducibility and accessibility (\href{https://github.com/bhavyaghai/WordBias}{github.com/bhavyaghai/WordBias}).
\section{Related Work}
\subsection{Bias in Word Embeddings}
The existing literature on bias in word embeddings can be broadly classified into bias identification and mitigation. For bias identification, a number of bias metrics are proposed like \textit{Subspace Projection}~\cite{bolukbasi2016man}, \textit{Relative Norm Difference}~\cite{100years}, \textit{Word Embedding Association Test (WEAT)}~\cite{narayan2017semantics}, etc., but there is no single agreed-upon method~\cite{group_words}. 
Our tool builds upon such bias identification metrics to explore the space of word associations and help detect biased associations. More specifically, our tool uses the \textit{Relative Norm Difference} metric as it is simple to interpret and can be easily extended for different kinds of biases. Previously, this metric has been used to capture biases against individual sensitive groups like females. In this work, we have used this metric to capture biases against intersectional groups as well. Once bias has been detected, there are a host of debiasing techniques which can be used for bias mitigation \cite{bolukbasi2016man, zhao2018gender, wang2020double}. However, we will not go into these details as our work is limited to bias discovery. 
Our work relates more closely with \citet{swinger2019biases}, who tries to find biases in word embeddings using purely algorithmic means compared to our visual analytics approach. Our dynamic visual interface makes the entire process more interactive and accessible to non-programmers. It also provides more flexibility by allowing the user to drive the bias discovery process as they see fit.

\subsection{Visual Tools}
Recent years have seen a spike in visual tools aimed at tackling Algorithmic fairness like Silva~\cite{silva}, FairVis~\cite{fairvis}, FairSight~\cite{fairsight}, What-If~\cite{what_if}, etc. All of these tools help detect Algorithmic Bias but they are mostly limited to tabular datasets. Moreover, many of these tools are designed to deal with individual biases and not intersectional biases. 
Our tool, WordBias, helps fill in this gap by helping discover intersectional biases encoded in word embeddings. Our tool relates closely to Google's Embedding Projector (GEP) ~\cite{smilkov2016embedding} which supports a custom projection adopted from \cite{bolukbasi2016man} to visualize bias. As a general-purpose tool primarily aimed at visualizing high dimensional data in 2D or 3D space, GEP has several limitations when it comes to exploring biases in word embeddings: (1) it does not support any bias quantification algorithm, (2) it is limited to visualizing only two types of bias simultaneously, and (3) its custom projection only allows one word to characterize a subgroup, say 'he' for males. In this work, we have tried to overcome all these limitations by carefully designing an interactive visual platform geared towards exploring social biases. 
\begin{figure*}
  \centering
  \includegraphics[width=0.90\textwidth]{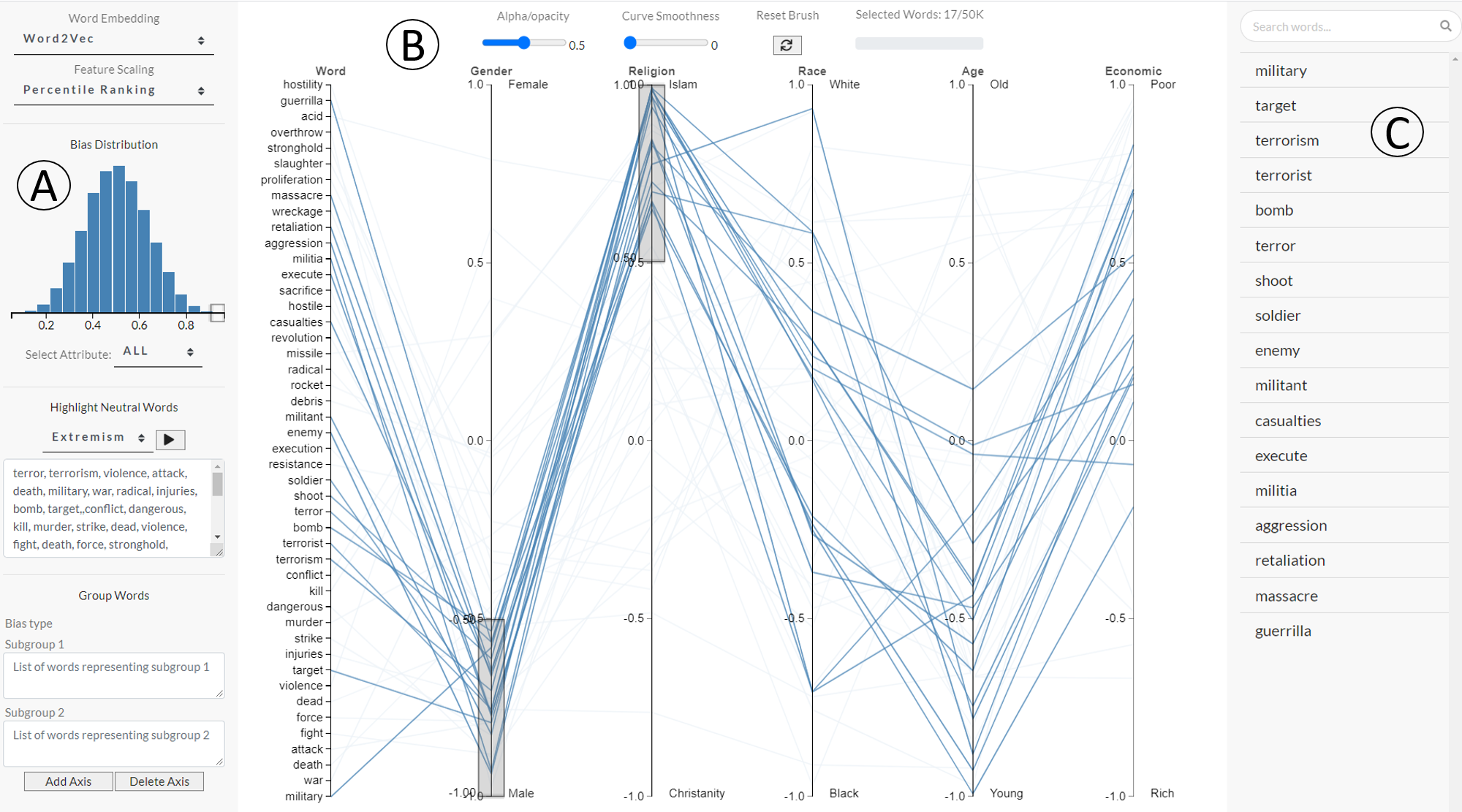}
  \caption{Visual interface of \textit{WordBias} using Word2Vec embedding. (A) The Control Panel provides options to select words to be projected on the parallel coordinates plot
  (B) The Main View shows the bias scores of selected words (polylines) along different bias types (axes)
  (C) The Search Panel enables users to search for a word and display the search/brushing results. 
  In the above figure, the user has brushed over 'Male' and 'Islam' subgroups. Words with strong association to both these subgroups are listed below the search box.}
 \label{fig:teaser}
 \Description[Visual interface of WordBias]{The image can be broken into 3 parts. (A) The Control Panel provides options to select words to be projected on the parallel coordinates plot
  (B) The Main View shows the bias scores of selected words (polylines) along different bias types (axes)
  (C) The Search Panel enables users to search for a word and display the search/brushing results. 
  In the above figure, the user has brushed over 'Male' and 'Islam' subgroups. Words with strong association to both these subgroups are listed below the search box. }
 \vspace{-1em}
\end{figure*}

\section{WordBias}

\subsection{Design Goals}
Based on the current literature and the problem at hand, we have identified the following four design goals:

\begin{itemize}
    
    \item [\textbf{G1.}] \textbf{Bias Scores}: Our tool should compute bias scores and accurately visualize them such that the user can quickly identify the different subgroups a word is associated to along with their degree of association. 

    \item [\textbf{G2.}] \textbf{Bias Exploration}: Our tool should support quick and intuitive exploration of words associated with a single subgroup say Males or an intersectional group say \textit{Rich White Females}.

     \item [\textbf{G3.}] \textbf{Bias Types}: The existing literature on biases in word embeddings is heavily skewed towards gender bias (93\%) followed by racial bias (54\%) \cite{rozado2020wide}. Our tool should support the exploration of these well known biases but also under-reported biases based on physical appearance, political leanings, etc. or any user-defined bias type.

    
    \item [\textbf{G4.}] \textbf{Data Volume}: Word embedding models can consist of millions of unique words. 
    Our tool should be designed to deal with a large volume of data at both the back and the front end to ensure a smooth user experience.
    
\end{itemize}

\subsection{Bias Quantification}
\label{sec:Bias_Quantification}
We have used the \textit{Relative Norm Difference} \cite{100years} to quantify the association (bias) of a word along different bias types. Like most bias metrics, it assumes that a given bias type, say gender, consists of two subgroups, say males and females. Each of such subgroups is defined using a set of words called \textit{group words}. For example, group words for males might include he, him, etc. while for females, it might include she, her, etc. Mathematically, a subgroup is expressed as the average of word embeddings for the words which define that subgroup. For a given bias type, let $\vec{g1}$, $\vec{g2}$ represent either subgroups. 
We then define the bias score for a word w with embedding $\vec{w}$ as follows:
\begin{equation}
\label{eq:1}
    Bias\_score(w) = cosine\_distance(\vec{w}, \vec{g1}) - cosine\_distance(\vec{w}, \vec{g2}) 
\end{equation}
A bias score can be understood as the association of a word toward a subgroup with respect to the other. The magnitude of the bias score represents the strength of the association and the sign indicates which subgroup it is associated to. We compute bias scores for each word across bias types using \autoref{eq:1} and then repeat this process for all words.  

\subsection{Feature Scaling}
Visualizing raw bias scores might be difficult to interpret and compare because the distribution of bias scores varies across bias types. For example, a 0.3 bias score for gender bias might mean a much stronger/weaker degree of association compared to the same score for race bias. To cope, WordBias supports two kinds of feature scaling methods namely, \textit{Min-Max Normalization} and \textit{Percentile Ranking}. Min-Max Normalization ensures that bias scores across bias types share the same range by simply stretching raw bias scores over the range [-1,1]. However, it is still difficult to compare bias scores because of the different standard deviation across bias types. 
To overcome this limitation, we use \textit{Percentile Ranking}. Each word is assigned a percentile score \cite{percentile_score} based on its ranking within its subgroup. For e.g., a 0.8 percentile score means that 80\% of all words associated with the same subgroup have a bias score less than or equal to the given word. This makes it easier to interpret and compare bias scores across different bias types (\textbf{G1, G2}). It should be noted that percentile scores can sometimes be misleading as they are not equally spaced. Lets say that raw bias scores for most words along a bias type is close to 0. However, we can still obtain high percentile scores for words which otherwise have negligible raw bias scores.   
Hence, we recommend trying both feature scaling methods to get a comprehensive picture.  

\subsection{Design Rationale}
The problem of visualizing biases against intersectional groups boils down to visualizing a large multivariate dataset where each word corresponds to a row and each column corresponds to a bias type. A straightforward solution for visualizing such high-dimensional data is to use standard \textit{dimensionality reduction} techniques like MDS, TSNE, biplot, etc. and then use popular visualization techniques like scatter plot. However, Algorithmic bias is a sensitive domain; we must make sure that we \textit{accurately} depict the biases of each word (\textbf{G1}). Hence, \textit{dimensionality reduction} and related techniques like the Data Context Map\cite{dataContextMap} are not an option because they almost always involve some information loss. Using such techniques might inflate/deflate real bias scores which might mislead the user. 

Next, we enumerated other possible ways to visualize multivariate dataset, like scatterplot matrix, radar chart, etc. and then started filtering these options based on the design challenges G1-G4. The scatter plot is a popular choice which is also used in Google's Embedding projector \cite{smilkov2016embedding}, but it is limited to three dimensions. A couple of more dimensions can be added by encoding radius and color of each dot yielding a plot that can visualize 5 dimensions; but such a plot will be virtually indecipherable. The scatterplot matrix can also be an option but it is more geared to visualizing binary relationships than the feature value of each point. Moreover, it becomes more space inefficient as the number of dimensions grow. Another alternative can be the biplot but it can be difficult to read and involves information loss. The radar plot provides for a succinct representation to visualize multivariate data but it can only handle a few points before polygons overlap and it becomes unreadable (defeating G4). 
We ended up with the parallel coordinate (PC) plot \cite{inselberg1990parallel} based on our design goals \textbf{G1-G4}. PC can visualize a significant number of points with multiple dimensions without any information loss (G1, G4). It also facilitates bias exploration and adding new bias types. To support plotting large numbers of points, we chose canvas over SVG and also used progressive rendering \cite{progressive_rendering} (\textbf{G4}). 

\begin{figure*}
\centering
\begin{minipage}{.30\textwidth}
  \centering
    \includegraphics[scale=0.50]{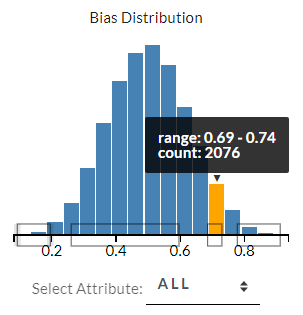}
  \captionof{figure}{Select words based on their bias scores by brushing on the x-axis of the histogram.}
  \label{fig:histogram}
  \Description[Histogram representing bias distribution]{On hovering over any bar in the histogram, its color changes and its corresponding bias score range along with the number of words appear as a tooltip.}
\end{minipage}%
\begin{minipage}{0.02\textwidth}
  \includegraphics[scale=0.05]{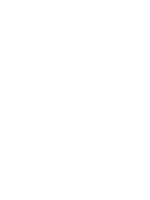}
\end{minipage}
\begin{minipage}{.58\textwidth}
  \centering
  \includegraphics[scale=0.23]{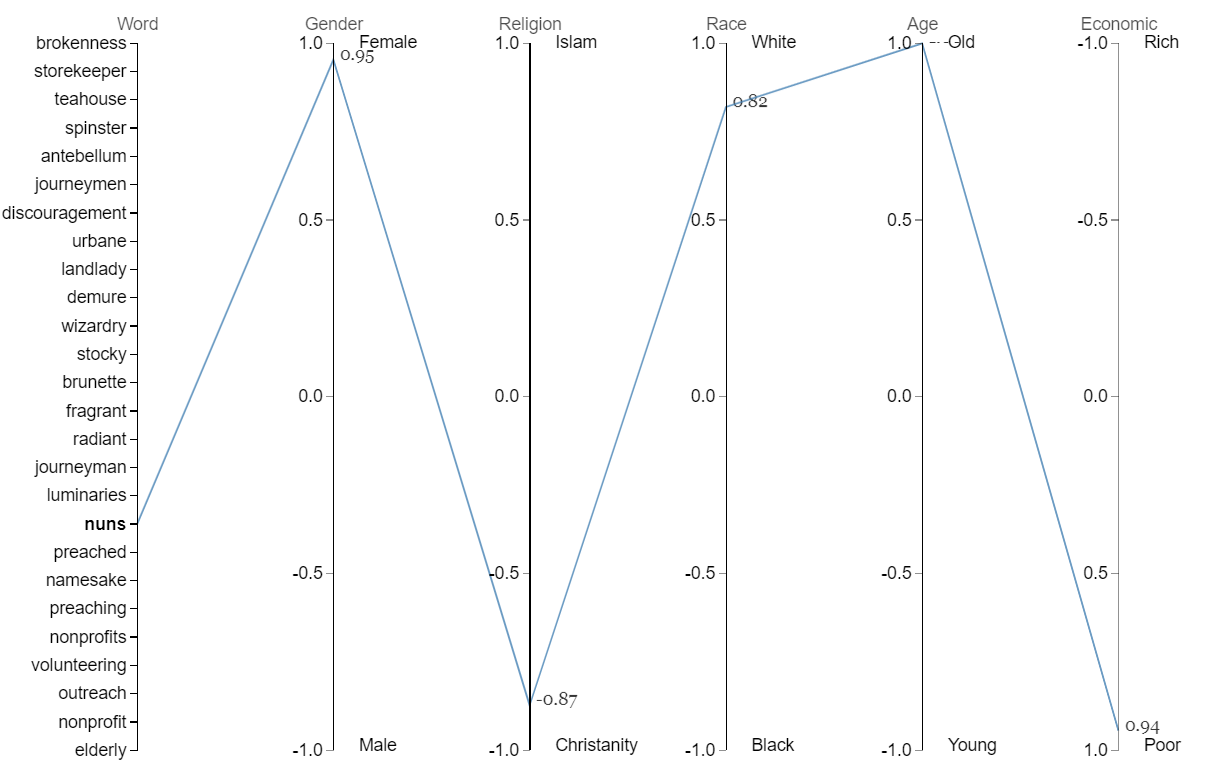}
  \captionof{figure}{On Hovering over the word 'nuns', 
  we can observe its association with 'Female', 'Christianity', 'White', 'Old' and 'Poor' subgroups.
  }
  \label{fig:hover}
  \Description[Parallel Coordinate]{On Hovering over the word 'nuns' in the parallel coordinate plot, its corresponding polyline gets highlighted. We can observe its association with 'Female', 'Christianity', 'White', 'Old' and 'Poor' subgroups.}
\end{minipage}
\vspace{-1em}
\end{figure*}

\subsection{Visual Interface}
\label{sec:vis_interface}
The visual interface can be classified into 3 components (see \autoref{fig:teaser}). Next, we will discuss each component in detail.  

\subsubsection{Main View}
At the very center is the Main View which consists of a parallel coordinate plot \cite{inselberg1990parallel} (see \autoref{fig:teaser} (B)). 
Each axis of the PC plot represents a type of bias based on gender, race, etc. and each piecewise linear curve, called \textit{polyline}, encodes a word. Either end of each axis represents a subgroup. For example, the gender axis encodes males and females on either extremes. The higher the magnitude of a word's bias score, the higher is the inclination of the corresponding polyline towards either group. We also have an additional axis, \textit{word}, which lists all the words currently displayed. On hovering over any word on the \textit{word} axis, its corresponding polyline gets highlighted (see \autoref{fig:hover}). This visualizes all different associations for the specific word (\textbf{G1}). On clicking over any word on the \textit{word} axis, the word and its synonyms get highlighted. Synonyms for a word are fetched via Thesaurus.com (using an API call) and from the nearest neighbors in the word embedding space.

To identify words with strong association toward a subgroup say females, the user can simply brush on the corresponding end of a given axis. Similarly, brushing either ends on multiple axes will help discover words associated to that intersectional group (\textbf{G2}). As shown in \autoref{fig:teaser}, the user may brush on \textit{Male} and \textit{Islam} ends on gender and religion axes to obtain words related to the intersectional group \textit{Muslim males}. Words selected via brushing are displayed under the search box (see \autoref{fig:teaser} (C)). 
At the top of the Main panel, there are sliders to customize \textit{Alpha/Opacity} and \textit{Curve smoothness} of the polylines. They are useful to see the underlying pattern between lots of polylines and to deal with the `crossing problem'\cite{graham2003using} respectively. \\


\subsubsection{Control Panel}
The left panel (see \autoref{fig:teaser} (A)), the Control Panel, allows the user to control what is displayed on the parallel coordinates plot. The user can choose the word embedding and the feature scaling method from the respective dropdown menus. 
It also contains a histogram accompanied by a dropdown menu. 
The dropdown menu contains a list of all bias types currently displayed in the parallel coordinates 
along with an \textit{ALL} option. 
The histogram serves two purposes. First, it helps users understand the underlying distribution of bias scores for the selected bias type across the word embedding.  
Second, it helps users deal with the problem of over-plotting by acting as a filtering mechanism (\textbf{G4}). 
The user can select single/multiple ranges of variable length on the x-axis of the histogram (see \autoref{fig:histogram}). The words whose bias score falls in the selected range(s) are displayed on the parallel coordinates. 
The \textit{ALL} option (default) paints an aggregate picture as it corresponds to the mean absolute bias score across all bias types. 


We precomputed commonly known biases based on gender, race, religion, age, etc. to jump-start the bias discovery process when the tool first loads (\textbf{G4}). We used group words from the existing literature for each bias type \cite{100years,narayan2017semantics,kozlowski2019geometry} (see Appendix). The user is free to investigate a new bias type or drop an existing one by using the Add/Delete axis button (\textbf{G3}). To add a new bias type say \textit{political orientation}, the user needs to fill in details like axis name, subgroup names and \textit{group words} under the `Group Words' section and click `Add Axis'. Here, the \textit{Group words} should be chosen carefully as they play a critical role in computing the bias scores. 

Lastly, we included a set of neutral words corresponding to categories like professions, personality traits, etc. which should ideally have no association with bias types like gender, race, etc. These words have been derived from existing literature \cite{100years, narayan2017semantics} (see Appendix). On clicking the `play' button, the currently selected set of neutral words will be highlighted in the PC plot. This provides a quick way to audit an embedding for potential biases.

\subsubsection{Search Panel}
The right panel (see \autoref{fig:teaser} (C)) enables a user to search for a specific word and see the respective search/brushing results. A user can simply lookup how a specific word associates with different groups by searching for it in the search box (\textbf{G1}). This will highlight the specific word and its synonyms in the parallel coordinates. The area under the search box is used to populate the list of synonyms and brushing results. 


\begin{table*}
  \caption{\textbf{Words with strong association with each intersectional group (within top 25 percentile of each constituting subgroup) in Word2vec embedding trained over Google News corpus. 
  }}
 \vspace{-1em}
  \label{tab:table}
  \centering%
  \begin{tabular}{p{0.25\linewidth}p{0.7\linewidth}}
    \toprule
   Intersectional Group & Associated Words \\
  \midrule
  Poor - Young - Black & disaster, struggle, tackle, chaos, woes, hunger, uprising, desperation, insecurity, rampage, roadblocks, scarcity, calamity, homophobia, shoddy, falter, jailbreak, mineworkers, marginalization, evictions \\
  
  Rich - Old - White & formal, attractive, appealing, desirable, castle, desserts, seaside, golfing, cordial, bungalow, fanciful, warmly, salty, nutty, gentler, aristocratic, snug, prim, urbane \\
  
  Black - Muslim - Male & gun, assassination, bullets, bribes, thugs, looted, dictators, electrocuted, cowards, agitating, storekeeper, looter, bleeping, lynch, strongman, disbelievers, hoodlums \\
  
  Young - Christian - Male & career, dominant, brilliant, lone, terrific, heroes, superb, epic, monster, prowess, heavyweights, excelled, superstars, supremacy, fearless, inexperience, mastery, crafty, ply, conquering, rampaging \\

Poor - Female & ostracism, brokenness, mortgages, eviction, brothels, witchcraft, traumatized, discrimination, layoffs, uninsured, sterilizations, abortion, powerlessness, sufferer, neediest, prostitution, microloans, distressed, homelessness, miscarry\\

White - Christian - Female & romantic, nuns, virgin, republicans, peachy, platonic, convent, radiant, unspoiled, unpersuasive, soppy, honeymooning, drippy, soapy\\

  \bottomrule
  \end{tabular}
\end{table*}

\section{Implementation}
WordBias is implemented as a web application built over python based web framework \textit{Flask}. On the back end, we used \textit{gensim} package to deal with word embeddings and \textit{PyThesaurus}\footnote[1]{\url{pypi.org/project/py-thesaurus}} to fetch synonyms from Thesaurus.com. For the front end, we used D3 based library \textit{Parallel Coordinates}\footnote[2]{github.com/syntagmatic/parallel-coordinates} and used \textit{D3.js}, \textit{Bootstrap}, \textit{noUiSlider}, etc. for rendering different visual components. 



\section{Case Study}
Let us assume a user, Divya (she/her), who works as a Data Scientist for a big Tech firm. 
Her team is tasked with building an automatic language translation tool. Made aware by the infamous \textit{Google Translate} example \cite{prates2019assessing,stanovsky2019evaluating}, she knows that such translation tools can be discriminatory toward minorities and can pose serious challenges for her organization. One of the ways in which bias can creep in is via word embeddings \cite{bios, bolukbasi2016man}. So, she needs to audit the word embedding for different social biases before using it. One way to explore/detect biases can be via purely algorithmic means i.e., writing custom program to test the different associations. Given that exploration is a dynamic process, so one might need to tweak and re-run the code repeatedly which can be tedious and cause delays. Moreover, analysing raw numbers for thousands of words across multiple bias types can be overwhelming and ineffective. Interactive visualization techniques excel at exploratory data analysis as they provide a faster, efficient and user friendly way to interact with massive datasets effectively \cite{keim2008visual}. So, Divya decides to use a visual analytics based tool, \textit{WordBias}, to audit her word embedding. Note that while we have used an embedding generated by word2vec \cite{mikolov2013distributed} trained over the Google News corpus, 
in a real world scenario this may be a word embedding trained over the company's private data. 

On first loading the tool, 
Divya observes that a small fraction of words are visualized which have strong association with multiple groups. 
These words correspond to the right tail of the histogram, i.e. they are words with high mean bias score.  
She hovers over some words like storekeeper, landlady, luminaries, nuns, etc. on the word axis to see their corresponding associations. Some of the associations are accurate and align well with her real world knowledge, like `landlady' and `nuns' have a strong association with females. In contrast, other associations, like `storekeeper' and `luminaries' have a strong male orientation which she views as problematic. It indicates that this word embedding might encode gender bias.  
To make sure that it is not a one-off case, 
she searches for the word `corrupt' in the Search panel. 
Just by looking at the parallel coordinates display, she can make out that the word `corrupt' and most of its synonyms like corruption, corrupted, crooked, unscrupulous, etc. have a strong association with Males and Blacks. 
This reaffirms the presence of gender bias and also indicates racial bias and bias against Black Males.

She carries on her investigation using different sets of words under the `Neutral Words' section in the control panel. 
Each time she finds a strong association of `ideally neutral' words with at least one kind of subgroup. When visualizing a set of \textit{Professions}, she finds words like teacher, nurse, dancer, etc. on brushing over the female subgroup and words like farmer, mechanic, physicist, laborer, etc. on brushing over the male subgroup. 
\autoref{fig:teaser} represents the case when she chooses to visualize words characterizing \textit{Extremism}. On brushing over the Male and Islam subgroup, she observes words like terrorist, bomb, aggression, etc. in the search panel. After this exercise, she is confirmed that this embedding encodes strong social biases against different groups as well as intersectional groups like Black males, Muslim males, etc. Her team might have to use different debiasing techniques before actually using this word embedding. 

The first step towards debiasing a word embedding is to identify the different impacted groups \cite{bolukbasi2016man, manzini2019black}. 
So, she explores different intersectional biases by selecting all the words using the histogram and then brushing over different subgroups. She finds lots of positive and negative stereotypes (biases) against multiple intersectional groups. Some of the more striking associations are described in \autoref{tab:table}. Overall, our tool helped Divya and her team to prevent a possible disaster by making them aware about the different social biases encoded in the word embedding. From here on, they can take multiple paths like trying to mitigate these biases, using a different word embedding, etc. They also need to be cautious about other possible sources of bias \cite{mehrabi2021survey} like training dataset to make sure that bias does not creep in.       

\section{Expert Evaluation}
We conducted a set of individual 45-60 min long semi-structured interviews with five domain experts. 
All experts E1-E5 are faculty members affiliated to departments like Computer Science (E1, E3), Sociology (E4, E5) and Business School (E2) at reputed R1 Universities. They have taught course(s) and/or published research paper(s) dealing with Algorithmic Fairness/Intersectionality.
Each expert was briefed about the problem statement and existing solutions. 
Thereafter, we demonstrated the different features, interactions and the workflow of our system using a case study.
Lastly, we solicited their comments on usability, utility, and scope for future improvements which are summarized as follows.

All experts found the interface to be \textit{intuitive} and \textit{easy to use}. Some experts found the interface to be a bit `overwhelming' at first glance. They were unsure of where to start interacting with the tool. 
However, a brief tutorial neutralized these concerns.
E4 commented, \textit{"Once you understand the tool, its very useful and you know what you are seeing"}. E3 commented that \textit{the UI looks clean} and \textit{actions required to accomplish tasks are simple and straightforward}. E2 commented, \textit{"Given a brief tutorial, most people should be able to get along quickly"}. 

On the utility front, E2 and E3 found this tool \textit{"Definitely useful"} for the NLP community while E1 stressed its utility for the Socio-Linguists and as an educational tool. E3 emphasized its \textit{broad} utility for developers, researchers and consumers, and also expressed interest in using this tool for teaching about bias in their NLP class. E4 emphasized the tool's utility for researchers and showed interest in loading their own custom word embedding into the tool. 
E4 added, \textit{"Anytime we want to ask a question from the data, we need to rerun the jupyter notebook which might take some time. This tool can cut down that Long feedback loop while providing rich information".}
E2 and E4 particularly liked that with WordBias users can dynamically add a new bias type on the go. This would make WordBias capable of supporting \textit{sentiment analysis} by encoding positive and negative sentiments on either extremes of an axis. 
Another important aspect of Wordbias which received appreciation is its \textit{accessibility} i.e., our tool can be hosted on a web server and then be easily accessed via a web browser without needing to install any software or dealing with github.   

For the future, most experts suggested to extend support for Contextualized word embeddings like BERT \cite{bert}, ELMo \cite{elmo}, etc. 
They pointed out that WordBias' current setup assumes a binary view of the real world since it only supports two subgroups per bias type. 
However, the real world is multi-polar.
They suggested to accommodate multiple subgroups like Whites, Blacks, Hispanics, Asians, etc. under a single bias type, say race.  
E1 highlighted that some of the bias variables like race and economic status might be correlated.  
Future work should account for such correlations while computing the bias scores. 
E5 suggested to encode multiple word embeddings representing different time periods on each axes. This will help in analysing how different biases evolve over time.   
E4 suggested to add a 'Download' button which can help store all words currently displayed in the tool along with their bias scores in CSV format.  

\section{Discussion, Limitations \& Future Work}

\paragraph{\textbf{Scalability}} We will discuss scalability on two aspects i.e. front-end rendering and back-end computation. On the frond end, we have used the parallel coordinate plot which can get cluttered as the number of points increases beyond a threshold. 
 We have used a number of visual analytics based techniques to ameliorate this issue, such as histogram based selection, changing opacity of lines, brushing, highlighting words on hover, etc. We have also used canvas based progressive rendering instead of SVG to render large data effectively (G4). Finally, there are also natural limitations on the number of bias types (axes) that can be differentiated in terms of their word associations.    

 Our current back-end can deal with words on a scale of $10^4$ while still maintaining a smooth user experience. As the number of words increases, the time for loading the word embedding and the time to calculate bias scores for a new bias type increases proportionally. Future work might use databases to store and query word embeddings to reduce load time. Furthermore, leveraging multiple compute cores will enable faster computation for any new bias type on the fly. 


\paragraph{\textbf{Quantifying Bias}} 
Measuring bias in word embeddings is an active research area and there is no consensus on a single best metric. In our case, we have used the \textit{Relative Norm Difference} metric. So, 
  the bias scores reported by our tool 
  are susceptible to the possible limitations of this metric and the group words used.
  The feature scaling methods, especially percentile ranking, can impact the perceived strength of an association. We recommend switching between different feature scaling methods (including raw bias scores) to get an accurate picture. 
Moreover, WordBias assumes a binary view of an inherently multi-polar world. This can impact the bias scores of words which do not fit into either categories. For eg., our tool reports white (race) orientation for the word `asian' even though its a different race altogether. One must interpret the bias scores responsibly in light of these limitations.
Future work might support multiple bias metrics 
to paint a more comprehensive picture and also include metrics which can better capture the multi-polar world.  

It is important to understand that the the term \textit{Intersectionality} has a broader meaning beyond multiplicity of identities \cite{crenshaw1989demarginalizing, crenshaw1990mapping, fleming2018less}. 
Quantifying such a complex sociological concept accurately needs more research. 
Our tool considers a narrow definition of Intersectionality where a word is linked to an intersectional group only if its relates strongly with each of the constituting subgroups. In reality, there can be cases like 'Hair Weaves' where a word is associated with an intersectional group (Black Females) even though it does not relate strongly with either constituting subgroups (Blacks or Females) \cite{ghavami2013intersectional}. Future work might incorporate bias metrics like EIBD \cite{guo2020detecting} which can capture such cases as well. 

\paragraph{\textbf{Utility}} Using a case study, we demonstrated how WordBias can be used as an \textit{auditing tool} by data scientists to probe for different kinds of social biases. 
Furthermore, the comments from the domain experts pointed at its possible utility for students and researchers. Given that WordBias does not require any programming expertise and can be easily accessed via a Web Browser, it can serve as an \textit{educational tool} for students and non-experts to learn how AI (word embedding model) might be plagued with multiple kinds of social biases. For researchers, our tool can expedite the bias discovery process by acting as a quick alternate to writing code. Future work might involve students, researchers and data scientists to further refine and evaluate the usability and utility of our tool for different target audiences.

\paragraph{\textbf{Group Words}} They play a critical role in computation of bias scores~\cite{group_words}. In our case, we have used group words which have been proposed in existing literature (see Appendix) to kick off the bias exploration process. The user is advised to examine the default set of group words and update them via the visual interface as required \cite{antoniak2021bad}. If the user chooses to add a new bias type (axis), they should choose the words carefully to get an accurate picture. So far, there is no objective way to choose group words. However, our tool can assist in selecting the most relevant group words by facilitating comparison against a set of alternatives (as recommended in \cite{antoniak2021bad}). Lets say the user wants to add a new axis for `political orientation' and they have multiple set of group words to choose from. In such a case, the user can add multiple axes corresponding to each set of group words. Thereafter, the user can explore and compare bias scores for different words across these axes. Group words corresponding to the axis which best aligns with the user's domain knowledge can be chosen.

\paragraph{\textbf{Word Embedding}} We have focused on static word embeddings trained on an English language corpus (word2vec). Similar social biases based on gender, etc. have been found in embeddings trained on other languages like French, Spanish, Hindi, German, Arabic, Dutch, etc. \cite{fr_es, hindi, german, arabic, dutch}. Furthermore, contextualized word embeddings like BERT \cite{bert}, Elmo \cite{elmo}, etc. have also been found to contain social biases based on gender, etc. \cite{wang2019gender, intersectional}. Future work will involve extending support for contextualized word embeddings and embeddings trained on other languages.

\section{Conclusion}
In this work, we designed, implemented and evaluated a novel visual interactive tool to discover intersectional biases in word embeddings. We demonstrated how our tool helped uncover biases against multiple intersectional groups encoded in Word2Vec embedding. The source of such biases can be training data, word embedding model or they might be false positives due to limitations of the bias metric or sub optimal group words. 
Future research might investigate the exact cause of such biases and develop effective counter strategies.   
\begin{acks}
We thank all domain experts for their time and feedback. This work was funded through NSF award IIS 1941613. 
\end{acks}

\bibliographystyle{ACM-Reference-Format}
\bibliography{references}

\appendix

\section{Preprocessing word embedding} 
Before loading the word embedding onto WordBias, we did some prepossessing similar to what is followed in the literature \cite{bolukbasi2016man}. We only considered words with all lower case alphabets and whose length is upto 20 characters long. We then sorted the resulting words by their frequency in the training corpus and picked the most frequent 50,000 words. We made sure to include group words like names, etc. if they don't make it in the final list.

\section{Feature Scaling}
\begin{figure}[h!]
  \begin{subfigure}[b]{0.3\textwidth}
    \includegraphics[width=\textwidth]{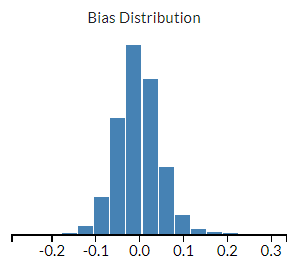}
    \caption{Raw bias scores}
  \end{subfigure}
  \hfill
  \begin{subfigure}[b]{0.3\textwidth}
    \includegraphics[width=\textwidth]{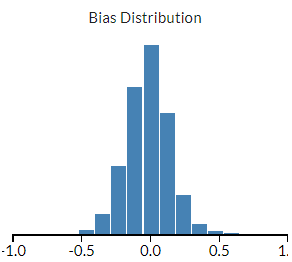}
    \caption{Min-Max Normalization}
  \end{subfigure}
  \hfill
  \begin{subfigure}[b]{0.3\textwidth}
    \includegraphics[width=\textwidth]{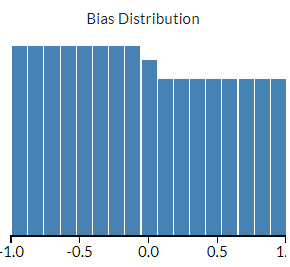}
    \caption{Percentile Ranking}
  \end{subfigure}
  \caption{Distribution of bias scores across 50k words in the Word2Vec Embedding.}
  \label{fig:feat_scaling}
\end{figure}

\begin{figure}[h!]
  \begin{subfigure}[b]{0.45\textwidth}
    \includegraphics[width=\textwidth]{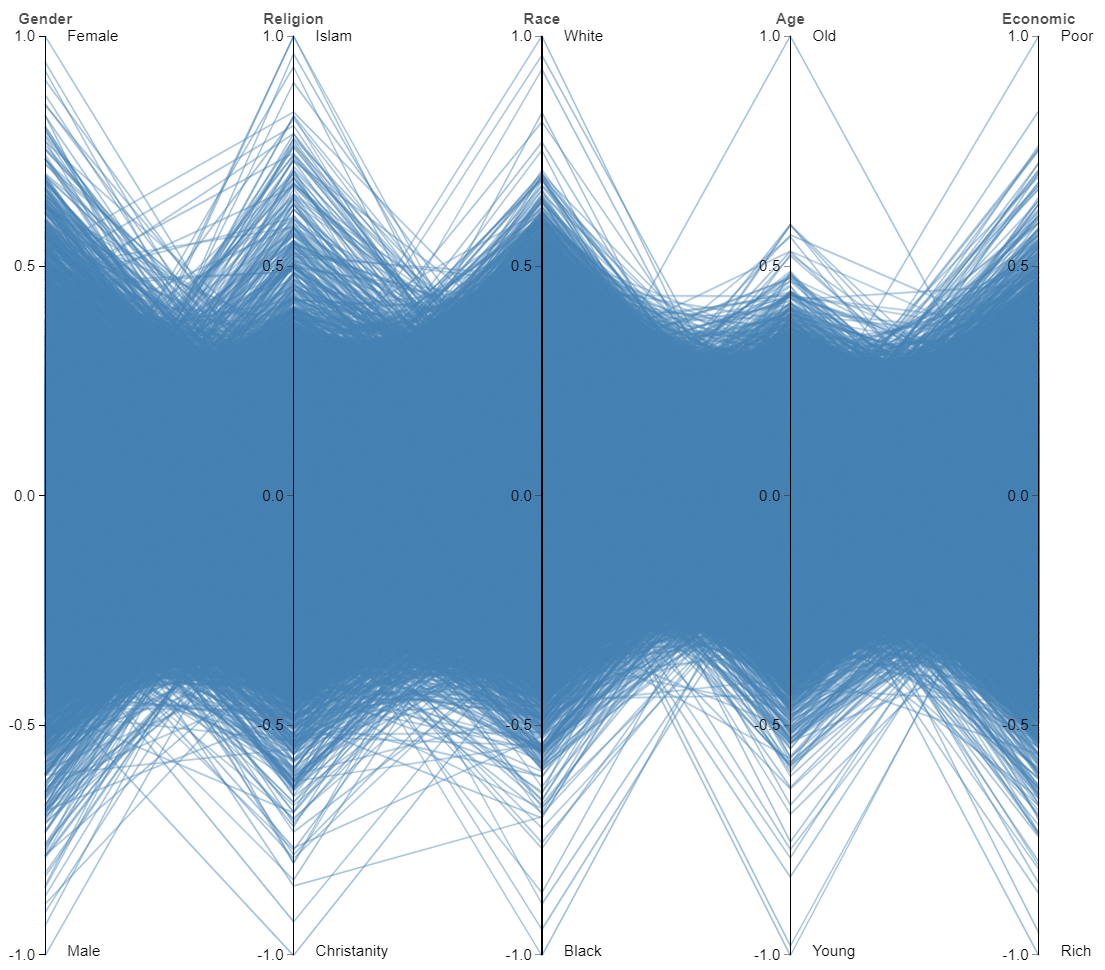}
    \caption{Min-Max Normalization}
  \end{subfigure}
  \hfill
  \begin{subfigure}[b]{0.45\textwidth}
    \includegraphics[width=\textwidth]{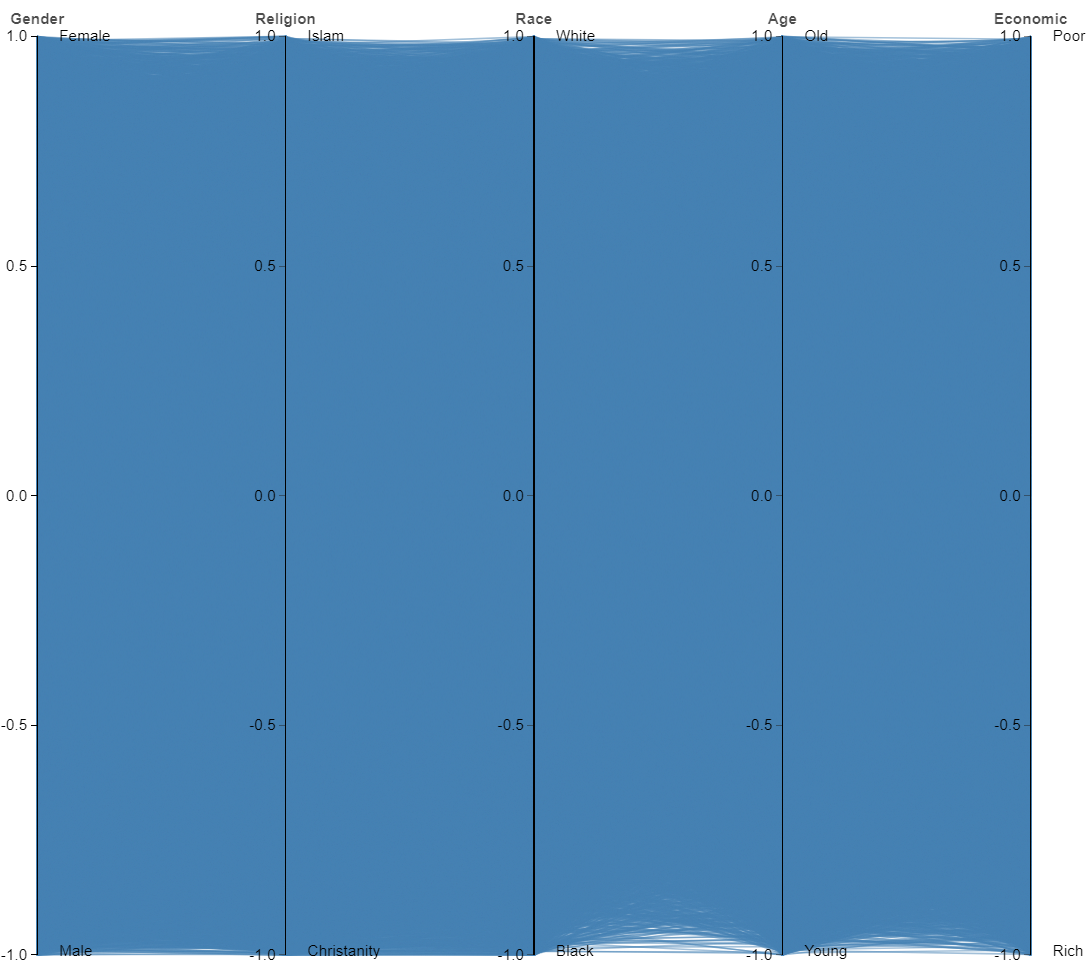}
    \caption{Percentile Ranking}
  \end{subfigure}
  \hfill
  \caption{Parallel coordinate plot for 50k words in the Word2Vec Embedding.}
  \label{fig:pc_feat_scaling}
\end{figure}

WordBias allows the user to choose between raw bias scores and two feature scaling methods namely, Min-Max Normalization and Percentile Ranking. Raw bias scores provides the most accurate representation but it can be a bit difficult to interpret. The other two feature scaling options makes the bias scores more comparable across bias types. \autoref{fig:feat_scaling} shows the distribution of mean bias scores for all 3 options. As we can see, the distribution of bias scores appear similar for (a) and (b) but their ranges on x-axis vary. This is because Min-Max normalization simply stretches the raw bias scores over the range [-1,1]. This figure also suggests that a large majority of words have small bias scores and only a few words on either ends have high bias scores. The distribution for Percentile ranking (\autoref{fig:feat_scaling} (c)) is quite different and interesting. It has the same range on x-axis [-1,1] as Min-Max normalization but the distribution of words across bias scores is much more uniform. We can observe the the bar length is different for bias scores greater than and less than 0. This is because we applied percentile ranking in a piece-wise fashion depending on the sign of the bias scores. \autoref{fig:pc_feat_scaling} further elucidates the difference in distribution of bias scores for Min-Max normalization and Percentile ranking.    

\section{Group Words}
\label{group_words}
By default, WordBias shows 5 kinds of biases namely Gender, Religion, Age, Race and Economic. Following are the list of words used to compute bias scores for each of those categories. These words are derived from existing literature \cite{kozlowski2019geometry, 100years, dev2019attenuating}. If any of these words aren't contained in the word embedding, they are ignored.    \\ 

\textbf{Male} (Gender) \cite{100years} \\
he, son, his, him, father, man, boy, himself, male, brother, sons, fathers, men, boys, males, brothers, uncle, uncles, nephew, nephews \\

\textbf{Female} (Gender) \cite{100years} \\ 
she, daughter, hers, her, mother, woman, girl, herself, female, sister, daughters, mothers, women, girls, sisters, aunt, aunts, niece, nieces \\

\textbf{Young} (Age) \cite{dev2019attenuating} \\ 
Taylor, Jamie, Daniel, Aubrey, Alison, Miranda, Jacob, Arthur, Aaron, Ethan \\

\textbf{Old} (Age) \cite{dev2019attenuating} \\
Ruth, William, Horace, Mary, Susie, Amy, John, Henry, Edward, Elizabeth  \\

\textbf{Islam} (Religion) \cite{100years} \\
allah, ramadan, turban, emir, salaam, sunni, koran, imam, sultan, prophet, veil, ayatollah, shiite, mosque, islam, sheik, muslim, muhammad \\

\textbf{Christainity} (Religion) \cite{100years} \\ 
baptism, messiah, catholicism, resurrection, christianity, salvation, protestant, gospel, trinity, jesus, christ, christian, cross, catholic, church \\

\textbf{Black} (Race) \cite{kozlowski2019geometry} \\
black, blacks, Black, Blacks, African, african, Afro  \\


\textbf{White} (Race) \cite{kozlowski2019geometry} \\
white, whites, White, Whites, Caucasian, caucasian, European, european, Anglo \\


\textbf{Rich} (economic) \cite{kozlowski2019geometry} \\
rich, richer, richest, affluence, advantaged, wealthy, costly, exorbitant, expensive, exquisite, extravagant, flush, invaluable, lavish, luxuriant, luxurious, luxury, moneyed, opulent, plush, precious, priceless, privileged, prosperous, classy \\

\textbf{Poor} (economic) \cite{kozlowski2019geometry} \\
poor, poorer, poorest, poverty, destitude, needy, impoverished, economical, inexpensive, ruined, cheap, penurious, underprivileged, penniless, valueless, penury, indigence, bankrupt, beggarly, moneyless, insolvent

\section{Neutral Words}
To quickly audit a given embedding for different biases, WordBias provides a set of words which should ideally be neutral for most kinds of biases like gender, race, etc. Following is the list of such neutral words based on different categories which are derived from existing literature \cite{100years, narayan2017semantics}. \\

\textbf{Profession} \\ 
teacher, author, mechanic, broker, baker, surveyor, laborer, surgeon, gardener, painter, dentist, janitor, athlete, manager, conductor, carpenter, housekeeper, secretary, economist, geologist, clerk, doctor, judge, physician, lawyer, artist, instructor, dancer, photographer, inspector, musician, soldier, librarian, professor, psychologist, nurse, sailor, accountant, architect, chemist, administrator, physicist, scientist, farmer \\

\textbf{Physical Appearance} \\ 
alluring, voluptuous, blushing, homely, plump, sensual, gorgeous, slim, bald, athletic, fashionable, stout, ugly, muscular, slender, feeble, handsome, healthy, attractive, fat, weak, thin, pretty, beautiful, strong \\

\textbf{Extremism} \\ 
terror, terrorism, violence, attack, death, military, war, radical, injuries, bomb, target,conflict, dangerous, kill, murder, strike, dead, violence, fight, death, force, stronghold, wreckage, aggression,slaughter, execute, overthrow, casualties, massacre, retaliation, proliferation, militia, hostility, debris, acid,execution, militant, rocket, guerrilla, sacrifice, enemy, soldier, terrorist, missile, hostile, revolution, resistance, shoot \\

\textbf{Personality Traits} \\
adventurous, helpful, affable, humble, capable, imaginative, charming, impartial, confident, independent, conscientious, keen, cultured, meticulous, dependable, observant, discreet, optimistic, persistent, encouraging, precise, exuberant, reliable, fair, trusting, fearless, valiant, gregarious, arrogant, rude, sarcastic, cowardly, dishonest, sneaky, stingy, impulsive, sullen, lazy, surly, malicious, obnoxious, unfriendly, picky, unruly, pompous, vulgar \\

\section{Few Examples}
In the following, we list a few words along with their associated subgroups as per WordBias. Here, we have chosen percentile ranking and considered an association significant if its corresponding bias score is $>= 0.5$. \\

\newpage
\textbf{(i) nazi} : Male - Christianity - White - Poor
\begin{figure}[H] 
  \centering 
  \includegraphics[width=0.8\columnwidth]{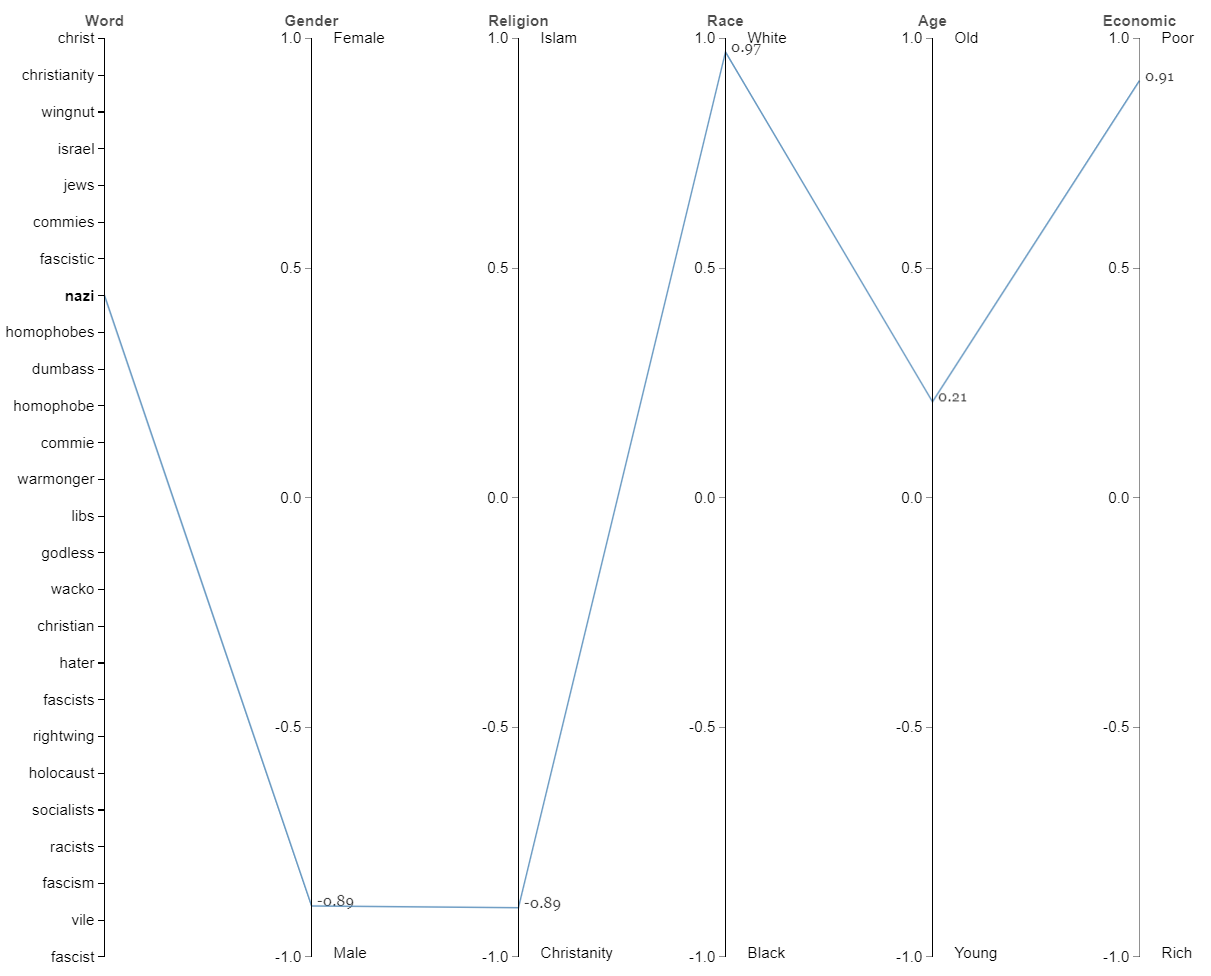}
\end{figure}

\textbf{(ii) beautiful} : Female - Christianity - Old - Rich
\begin{figure}[H] 
  \centering 
  \includegraphics[width=0.8\columnwidth]{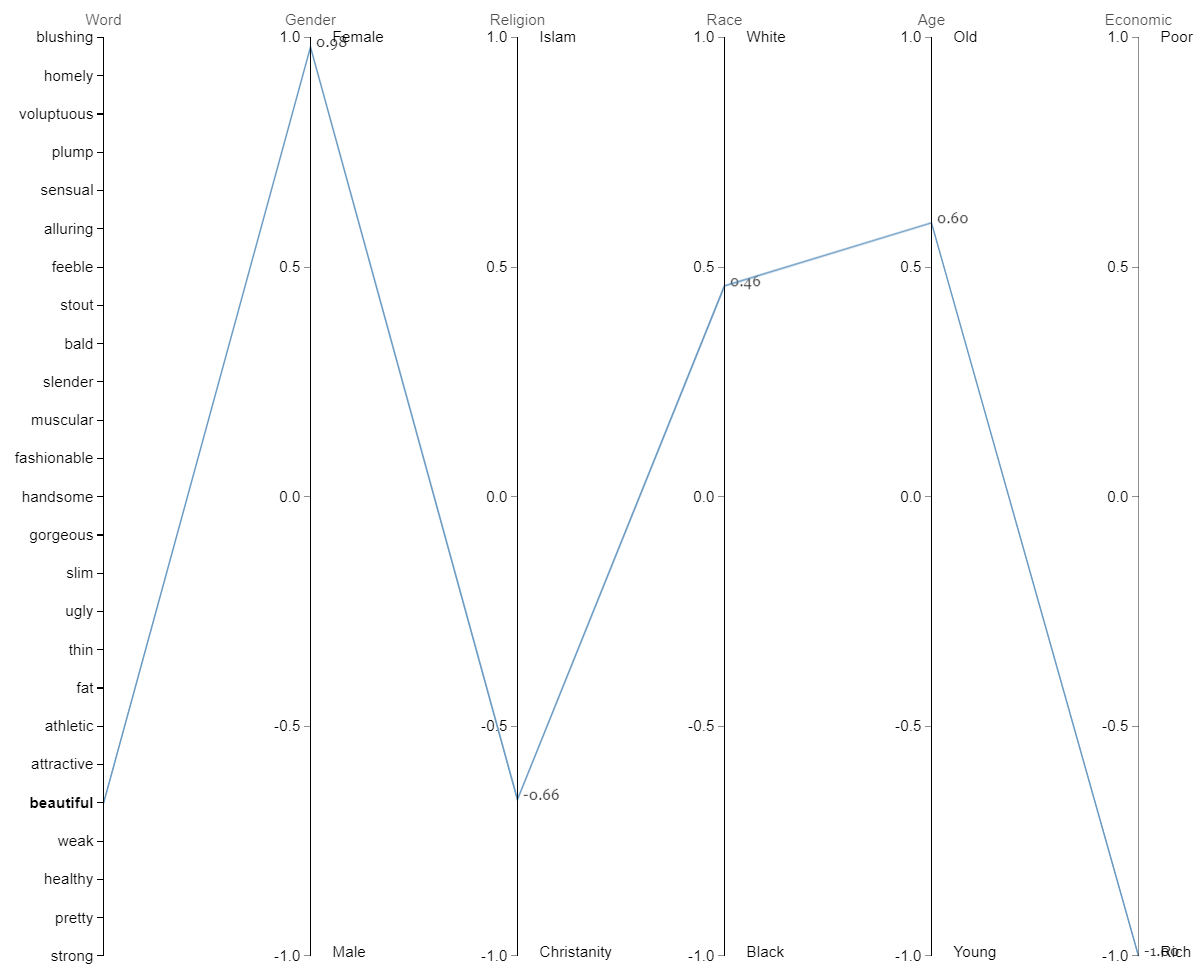}
\end{figure}

\newpage
\textbf{(iii) pretty} : Christianity - White - Young
\begin{figure}[H] 
  \centering 
  \includegraphics[width=0.8\columnwidth]{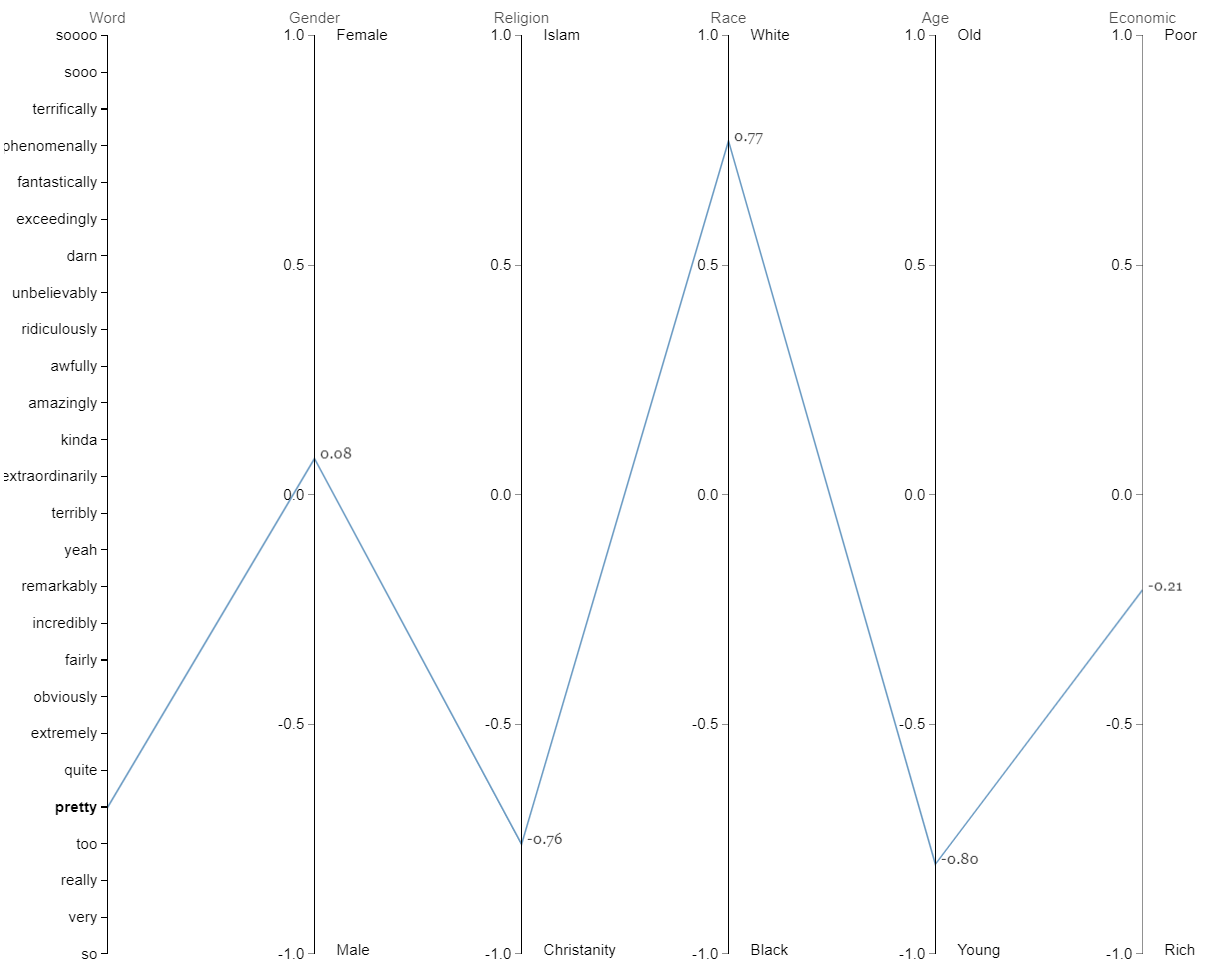}
\end{figure}

\textbf{(iv) homicides} : Female - Black - Poor
\begin{figure}[H] 
  \centering 
  \includegraphics[width=0.8\columnwidth]{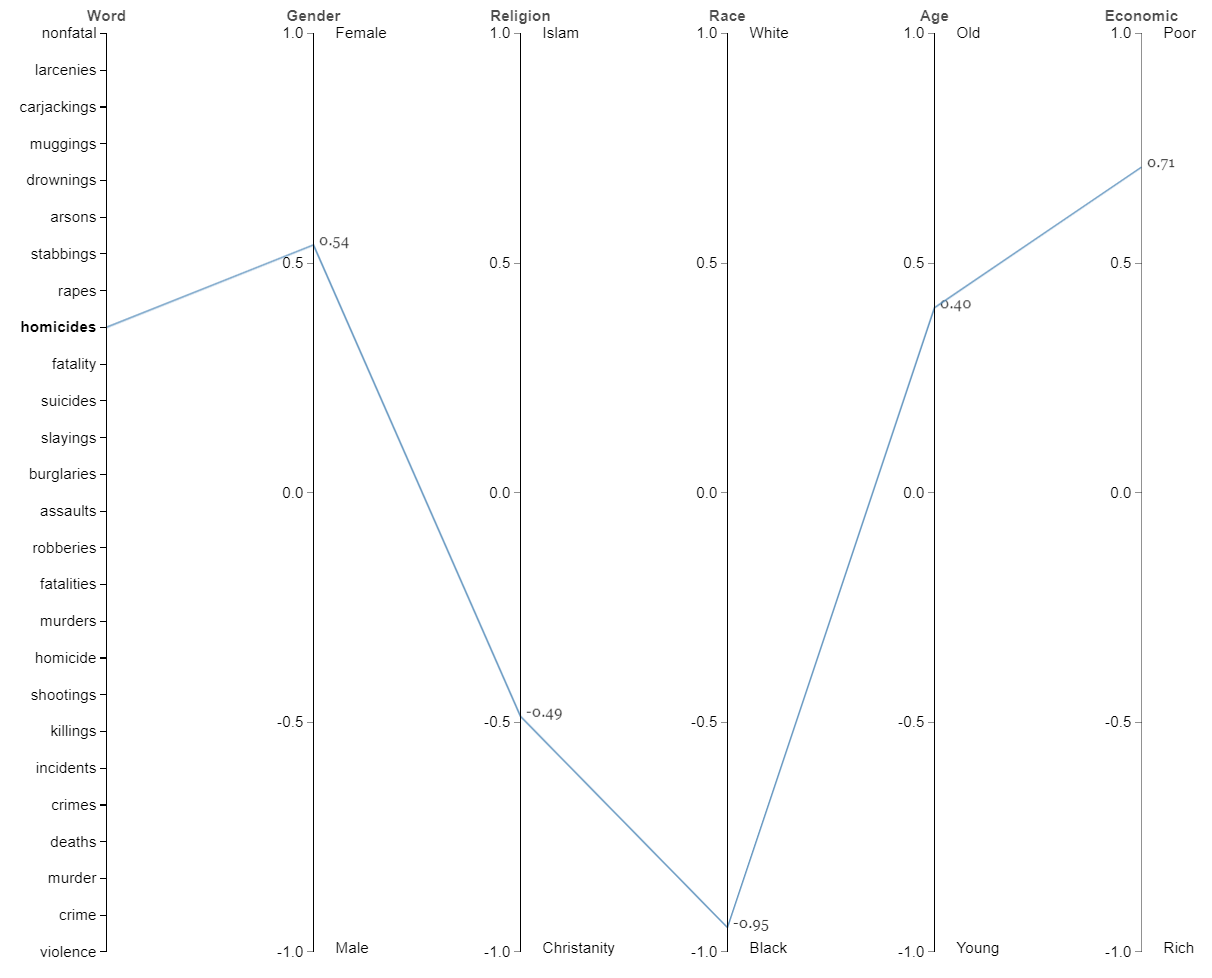}
\end{figure}

\newpage
\textbf{(v) picky} : Female - White - Young - Rich
\begin{figure}[H] 
  \centering 
  \includegraphics[width=0.8\columnwidth]{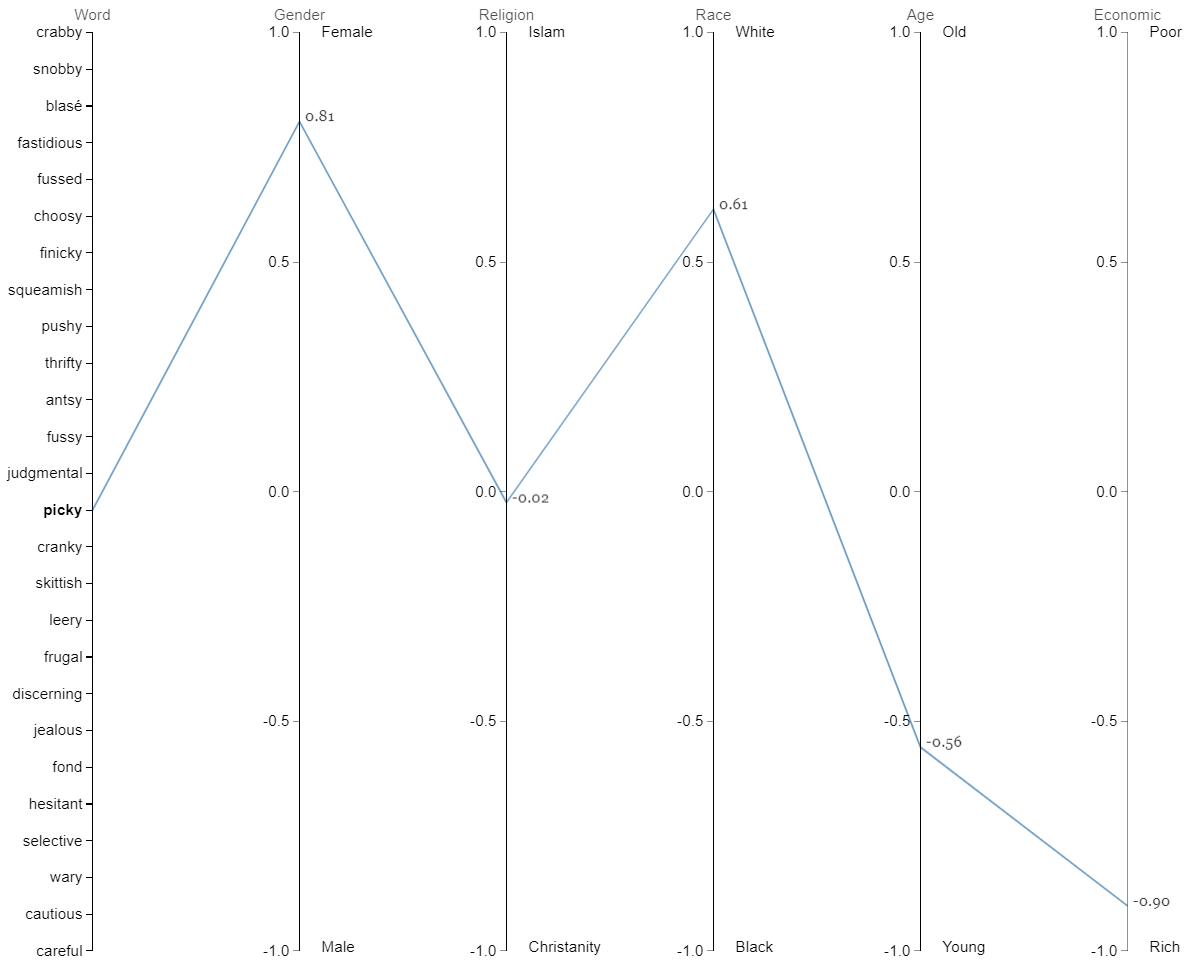}
\end{figure}

\textbf{(vi) terror} : Male - Islam - Young
\begin{figure}[H] 
  \centering 
  \includegraphics[width=0.8\columnwidth]{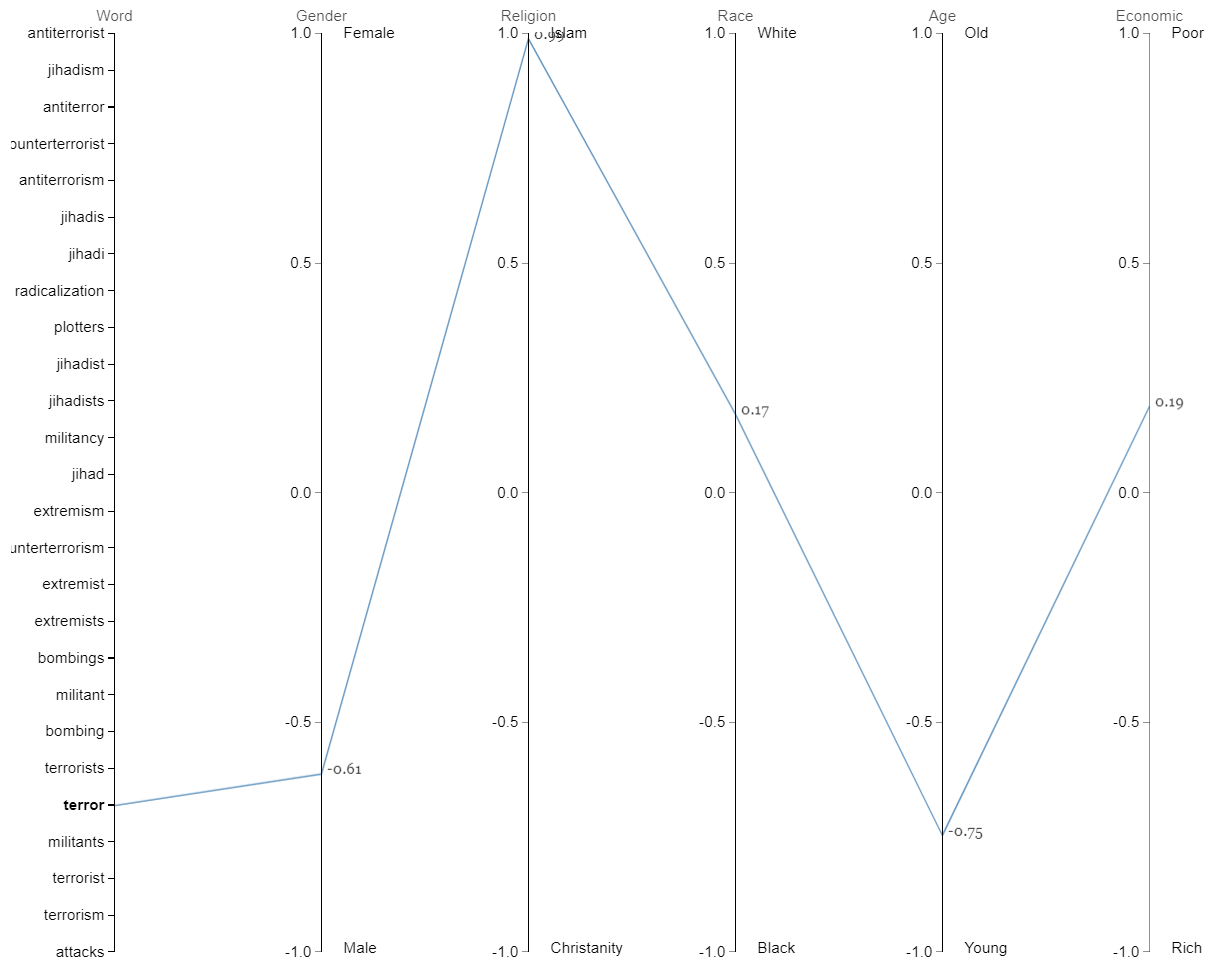}
\end{figure}

\newpage
\textbf{(vii) prostitute} : Female - Poor
\begin{figure}[H] 
  \centering 
  \includegraphics[width=0.8\columnwidth]{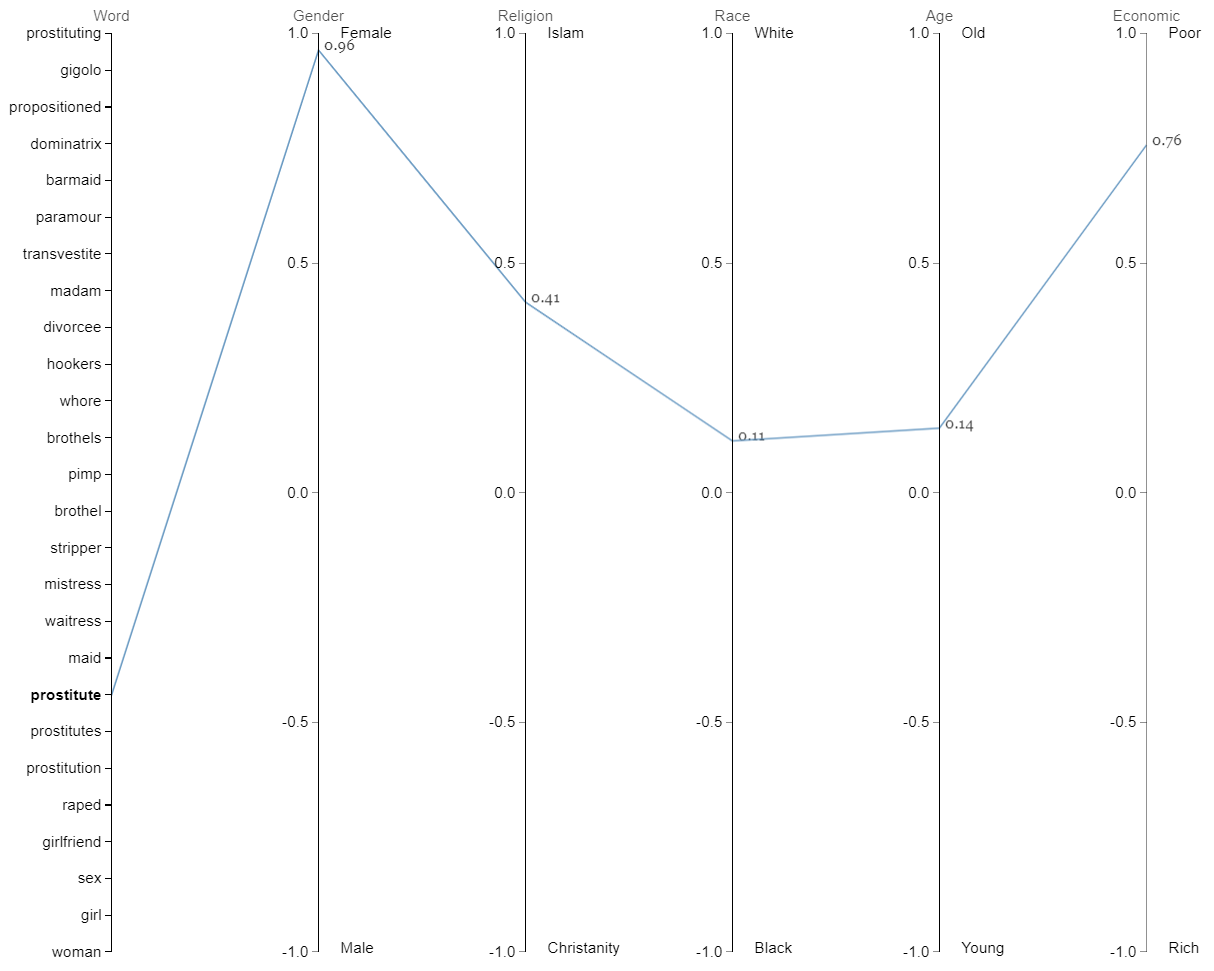}
\end{figure}

\textbf{(viii) clever} : Male - Christianity - Young - Rich
\begin{figure}[H] 
  \centering 
  \includegraphics[width=0.8\columnwidth]{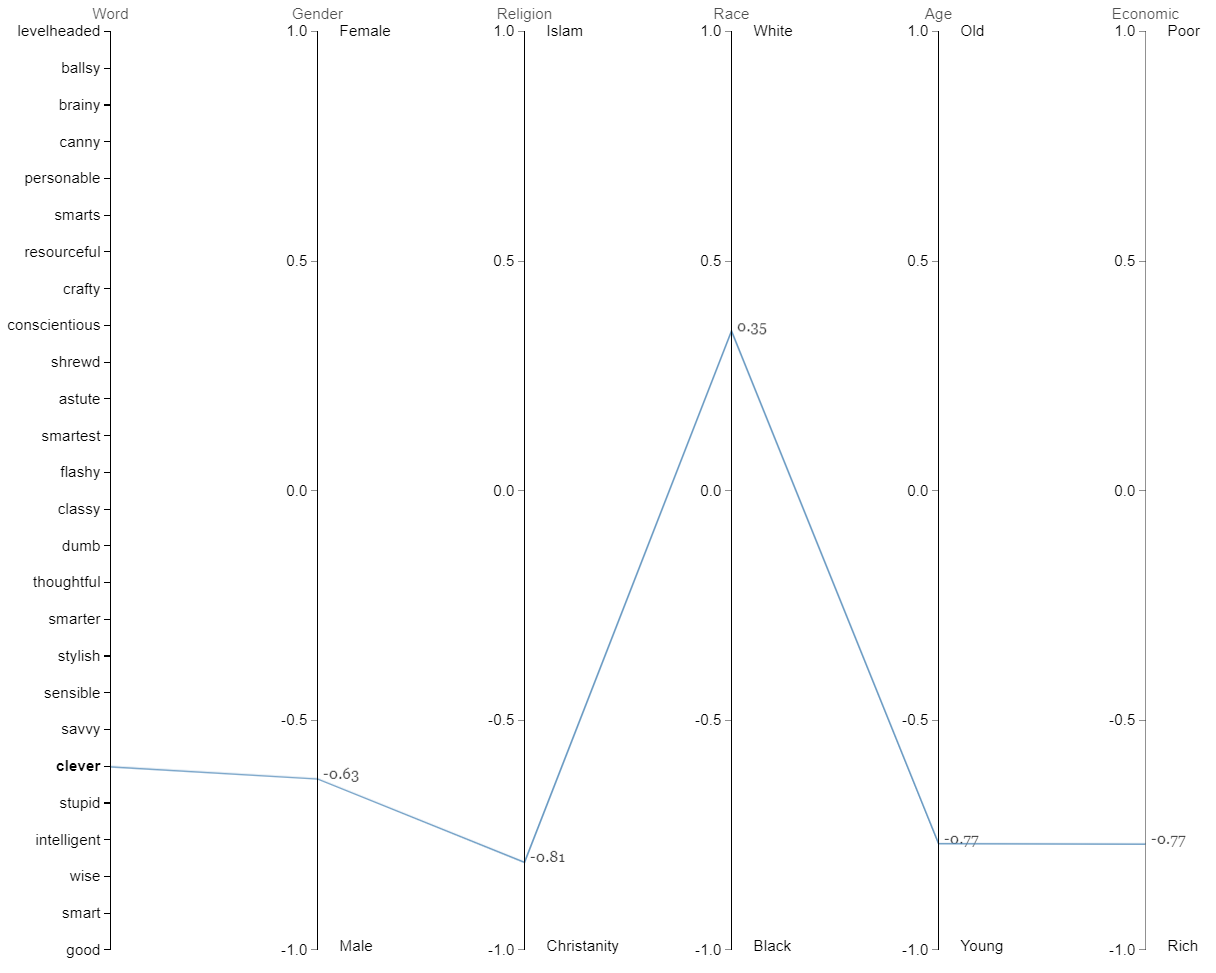}
\end{figure}

\newpage
\textbf{(ix) dictator} : Male - Islam - Black - Old - Poor
\begin{figure}[H] 
  \centering 
  \includegraphics[width=0.8\columnwidth]{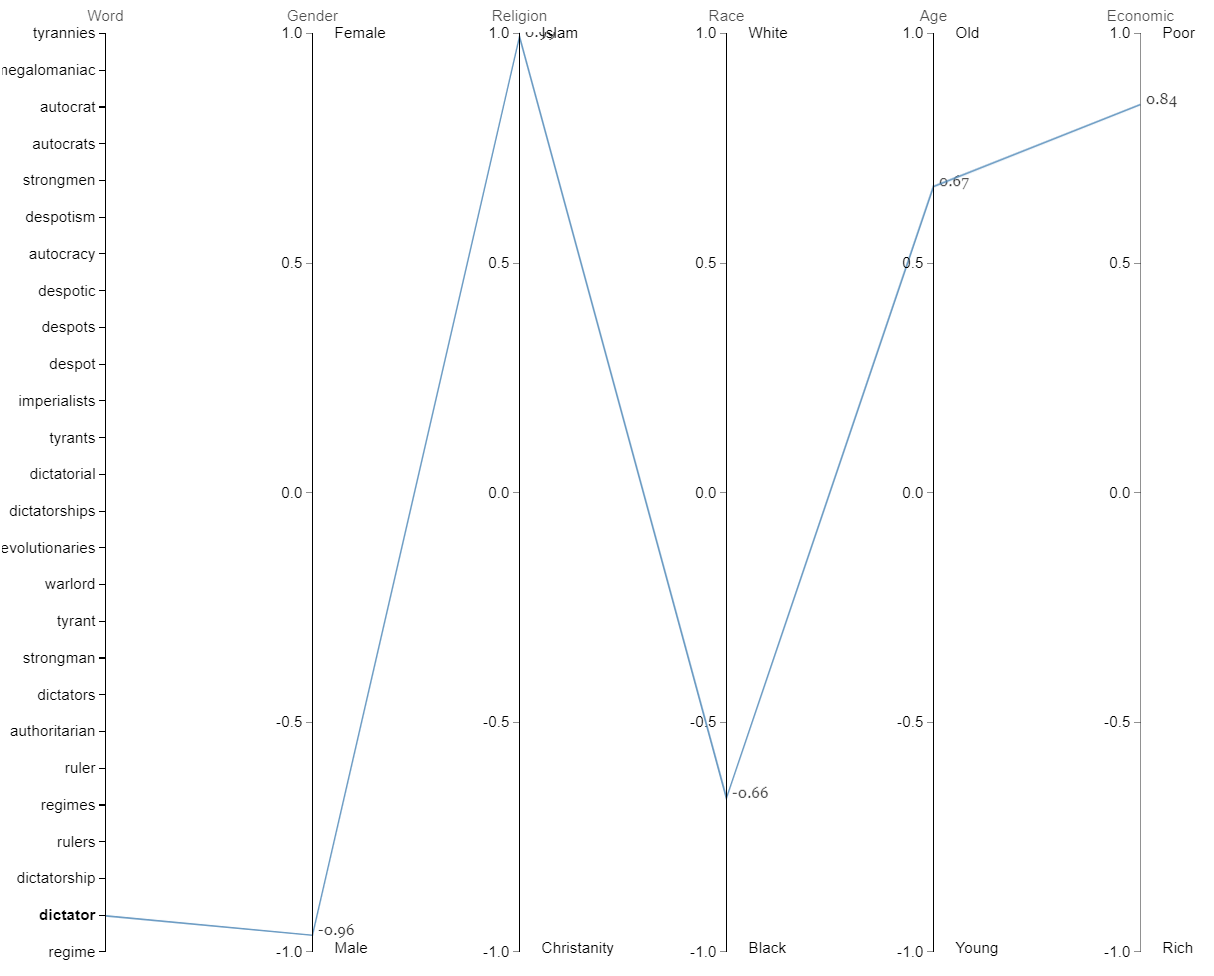}
\end{figure}

\textbf{(x) janitor} : Male - Old - Poor
\begin{figure}[H] 
  \centering 
  \includegraphics[width=0.8\columnwidth]{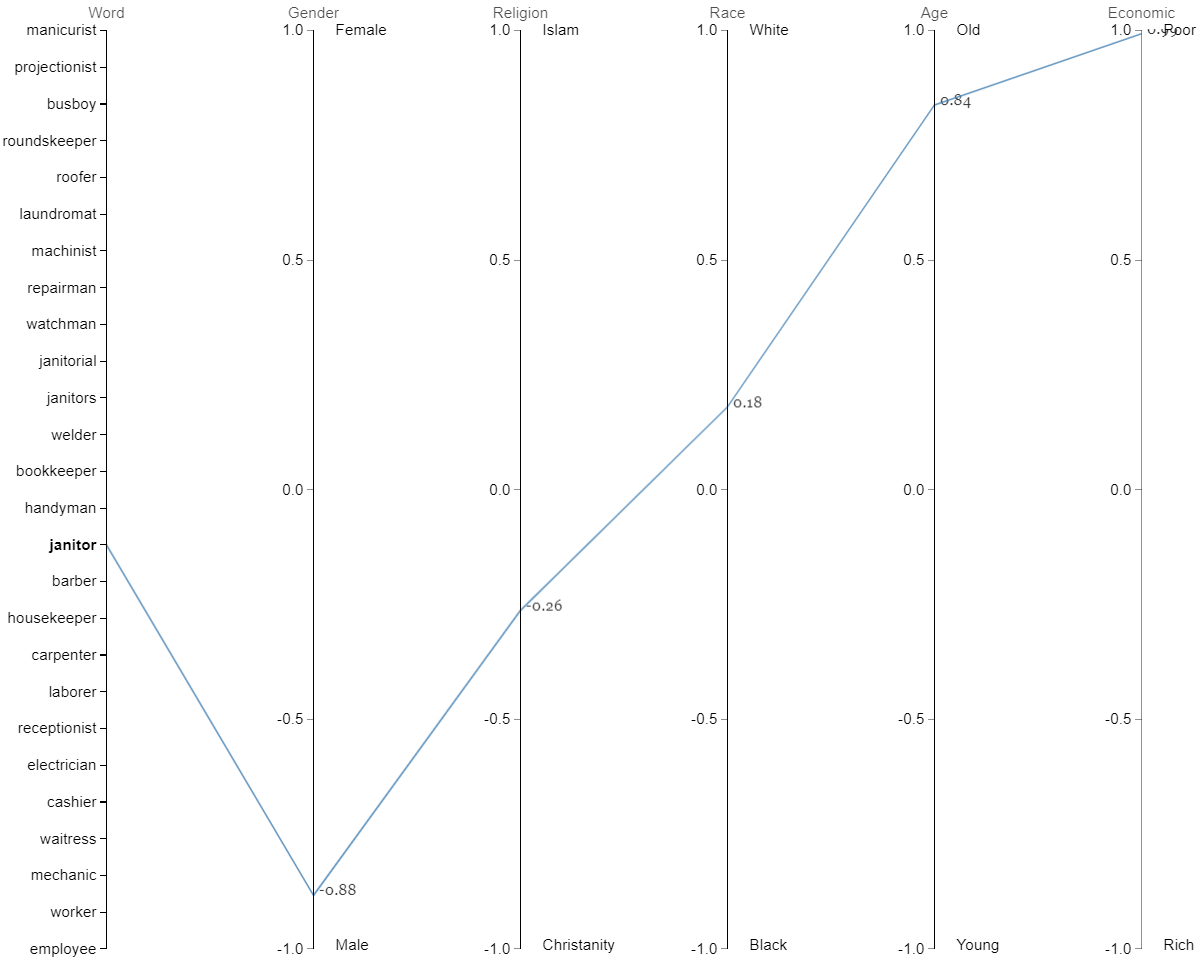}
\end{figure}

\newpage
\textbf{(xi) militia} : Male - Islam - Black - Poor
\begin{figure}[H] 
  \centering 
  \includegraphics[width=0.8\columnwidth]{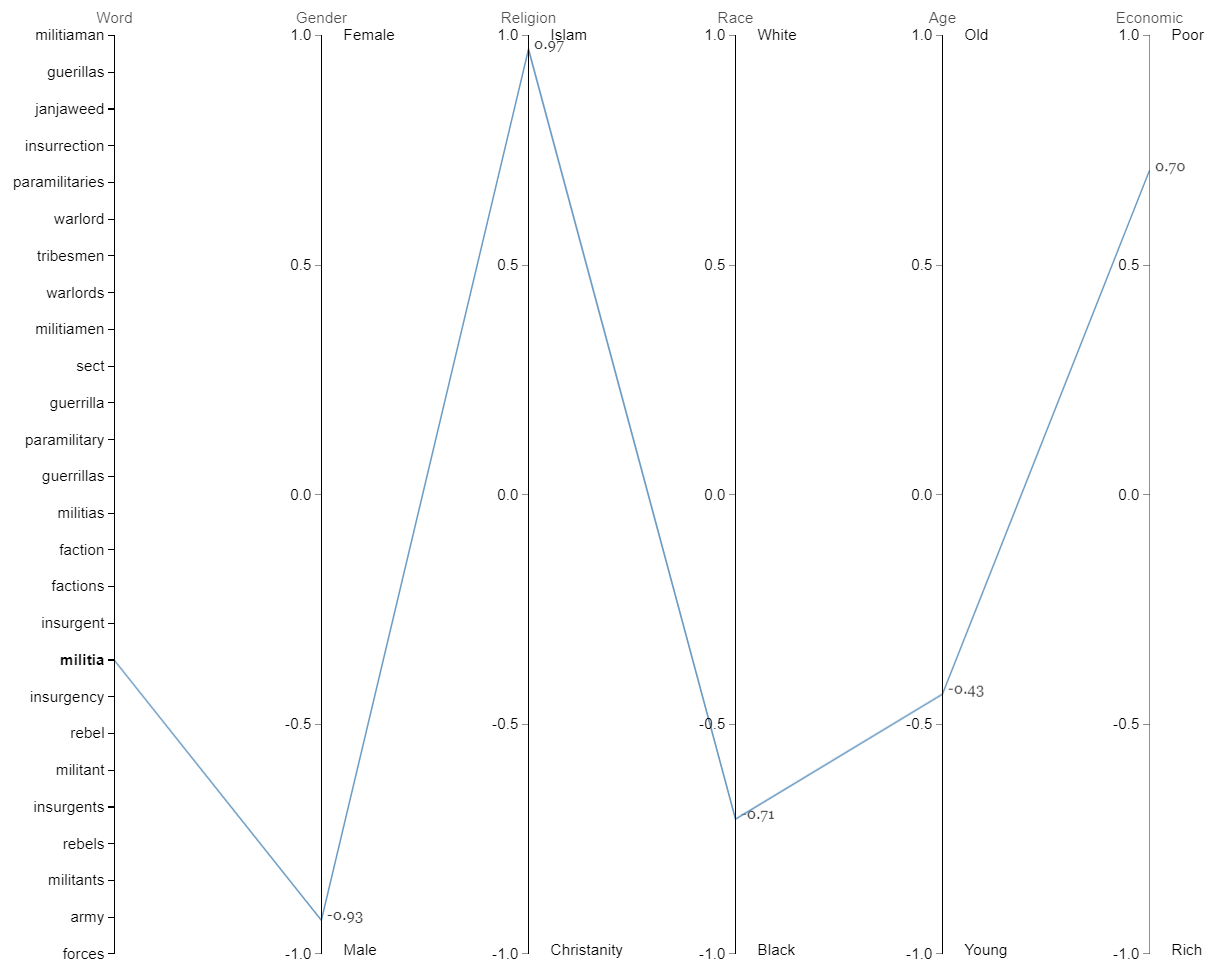}
\end{figure}


\end{document}